\def\year{2022}\relax
\documentclass[letterpaper]{article} 
\usepackage{aaai22}  
\usepackage{times}  
\usepackage{helvet}  
\usepackage{courier}  
\usepackage[hyphens]{url}  
\usepackage{graphicx} 
\usepackage{natbib}  
\usepackage{caption} 
\DeclareCaptionStyle{ruled}{labelfont=normalfont,labelsep=colon,strut=off} 
\usepackage{algorithm}
\usepackage{algorithmic}
\usepackage{bbm}
\usepackage{subcaption}
\usepackage[switch]{lineno}
\usepackage{color}

\usepackage{booktabs}
\usepackage{comment}
\usepackage{amsmath,amssymb}
\usepackage{diagbox}
\usepackage{multirow}
\usepackage{makecell}
\usepackage{bbding}

\usepackage{newfloat}
\usepackage{listings}
\floatstyle{ruled}
\newfloat{listing}{tb}{lst}{}
\floatname{listing}{Listing}
\title{GenCo: Generative Co-training for Generative Adversarial Networks with Limited Data}
\author {
    Kaiwen Cui\thanks{ indicates equal contribution.},
    Jiaxing Huang\footnotemark[1],
    Zhipeng Luo,
    Gongjie Zhang,
    Fangneng Zhan,
    Shijian Lu\thanks{ corresponding author.}
}
\affiliations {
    School of Computer Science Engineering, Nanyang Technological University\\
    \{kaiwen001, zhipeng001\}@e.ntu.edu.sg,  \{jiaxing.huang, Gongjiezhang, fnzhan, shijian.lu\}@ntu.edu.sg
}

\usepackage{bibentry}

\begin{document}
\maketitle

\begin{abstract}
Training effective Generative Adversarial Networks (GANs) requires large amounts of training data, without which the trained models are usually sub-optimal with discriminator over-fitting. Several prior studies address this issue by expanding the distribution of the limited training data via massive and hand-crafted data augmentation. We handle data-limited image generation from a very different perspective. Specifically, we design GenCo, a Generative Co-training network that mitigates the discriminator over-fitting issue by introducing multiple complementary discriminators that provide diverse supervision from multiple distinctive views in training. We instantiate the idea of GenCo in two ways. The first way is Weight-Discrepancy Co-training (WeCo) which co-trains multiple distinctive discriminators by diversifying their parameters. The second way is Data-Discrepancy Co-training (DaCo) which achieves co-training by feeding discriminators with different views of the input images  ($e.g.$, different frequency components of the input images). Extensive experiments over multiple benchmarks show that GenCo achieves superior generation with limited training data. In addition, GenCo also complements the augmentation approach with consistent and clear performance gains when combined. 
\end{abstract}

\section{Introduction} \label{sec Intro}

Generative Adversarial Networks (GANs)~\cite{goodfellow2014generative} have achieved great successes in various image generation tasks such as image-to-image translation~\cite{zhu2017unpaired, isola2017image, park2019semantic}, domain adaptation~\cite{hoffman2018cycada,luo2019taking, hsu2020progressive}, super resolution~\cite{ledig2017photo, wang2018esrgan,wang2018recovering} and image in-painting~\cite{yu2018generative,yu2019free, nazeri2019edgeconnect}. Nevertheless, high-fidelity image generation usually requires large amounts of training samples which are laborious and time-consuming to collect. Data-limited image generation, which aims to generate realistic and high-fidelity images with a small number of training samples, is a very meaningful yet challenging task in image generation.

\begin{figure}[!t]
\centering
\includegraphics[width=1.0\linewidth]{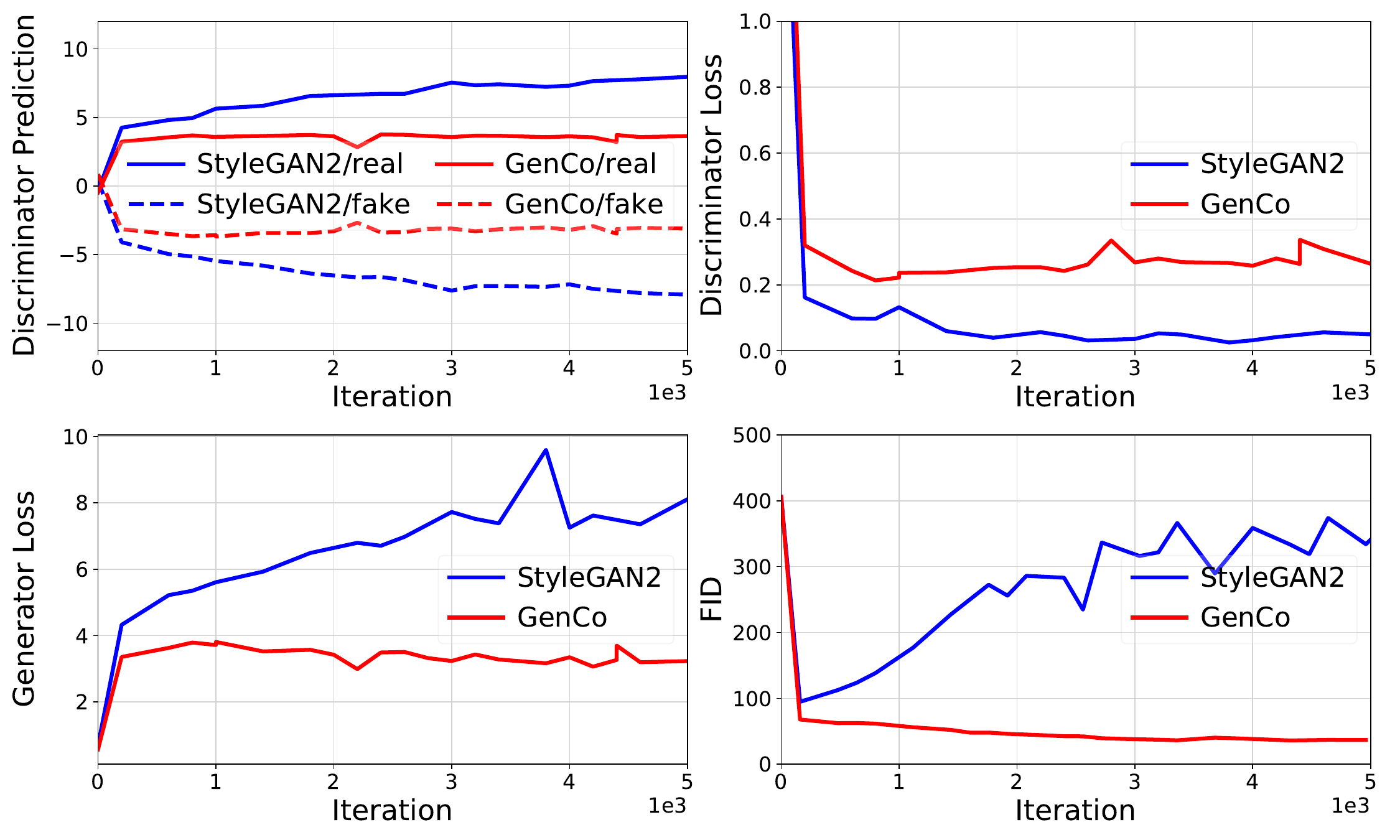}
\caption{
The proposed GenCo improves data-limited image generation clearly (on 100-shot-Obama dataset): With limited training samples, discriminator in most GANs such as StyleGAN2 tends to become over-fitting and produces very high-confidence scores and very small 
discriminator loss as shown in the two upper graphs. The very small discriminator loss further leads to very large generator loss as well as gradients which cause 
training to diverge and generation to deteriorate 
as shown in the two lower graphs. The proposed GenCo mitigates the discriminator over-fitting effectively with more stable training and better generation.}
\label{fig1}
\end{figure}

With limited training samples, the trained generation model suffers from discriminator over-fitting~\cite{zhao2020differentiable, karras2020training, tseng2021regularizing} which leads to degraded generation. Specifically, over-fitting discriminator produces very high prediction scores and very small  
discriminator loss as illustrated in the two upper graphs in Fig. \ref{fig1}. The very small discriminator loss then leads to very large generator loss and gradients which accumulate during training and lead to training divergence and degraded generation~\cite{pascanu2012understanding, pascanu2013difficulty} as illustrated in the two lower graphs in Fig. \ref{fig1}.
The over-fitting issue has attracted increasing interest recently, and the prevalent approach addresses the issue through massive data augmentation. The idea is to massively augment the limited training samples to expand the data distributions as much as possible. 
Though prior studies~\cite{karras2020training, zhao2020differentiable} demonstrate the effectiveness of this approach, they address the problem at the input end only without considering much about features and models.
In addition, some work \cite{tseng2021regularizing} addresses the over-fitting issue by 
including certain regularization into the discriminator loss.

We tackle the over-fitting issue from a very different perspective. 
Specifically, 
we introduce the idea of co-training into the data-limited image generation task, aiming to learn limited data from multiple distinctive yet complementary views. Co-training was originally proposed to boost the inspection performance when only limited data is available~\cite{blum1998combining}. It alleviates the data constraint effectively by employing multiple classifiers that learn from different views and capture complementary information about the limited data. In recent years, co-training has been adopted in different deep learning tasks such as semi-supervised image recognition~\cite{qiao2018deep}, unsupervised domain adaptation~\cite{saito2018maximum,luo2019taking}, etc., where the amount of training data becomes more critical as compared with traditional learning tasks without using deep neural networks.

Specifically, we design GenCo, a Generative Co-training network that adapts the idea of co-training into data-limited image generation for tackling its inherent over-fitting issue. GenCo trains the generator with multiple discriminators that mitigate the over-fitting issue by learning from multiple distinct yet complementary views of the limited data.
We design two instances of GenCo that enable the discriminators to learn from distinctive and comprehensive views.
 The first is Weight-Discrepancy Co-training (WeCo) which co-trains multiple distinctive discriminators by diversifying their parameters with a weight discrepancy loss. The second is Data-Discrepancy Co-training (DaCo) that co-trains distinctive discriminators by feeding them with different views of the input images. In our design, we extract different frequency components of each training image to form different views. The proposed GenCo mitigates the discriminator over-fitting issue and improves data-limited image generation effectively as illustrated in Fig. \ref{fig1}, more details to be discussed in the Experiments section.

The contribution of this work can be summarized in three aspects. \textit{First}, we propose to tackle the data-limited image generation challenge from a co-training perspective. To this end, we design GenCo, a Generative Co-training network that mitigates the discriminator over-fitting issue effectively by training the generator with multiple distinctive discriminators. \textit{Second}, we design two instances of GenCo that are complementary to each other, namely, WeCo that introduces weight discrepancy loss to diversify multiple discriminators and DaCo that learns distinctive discriminators by employing different views of input images. \textit{Third}, extensive experiments show that GenCo achieves superior generation quality and it is also complementary with the state-of-the-art augmentation approach with consistent performance gains.

\begin{figure*}[t!] 
\centering
  \includegraphics[width=1.0\linewidth]{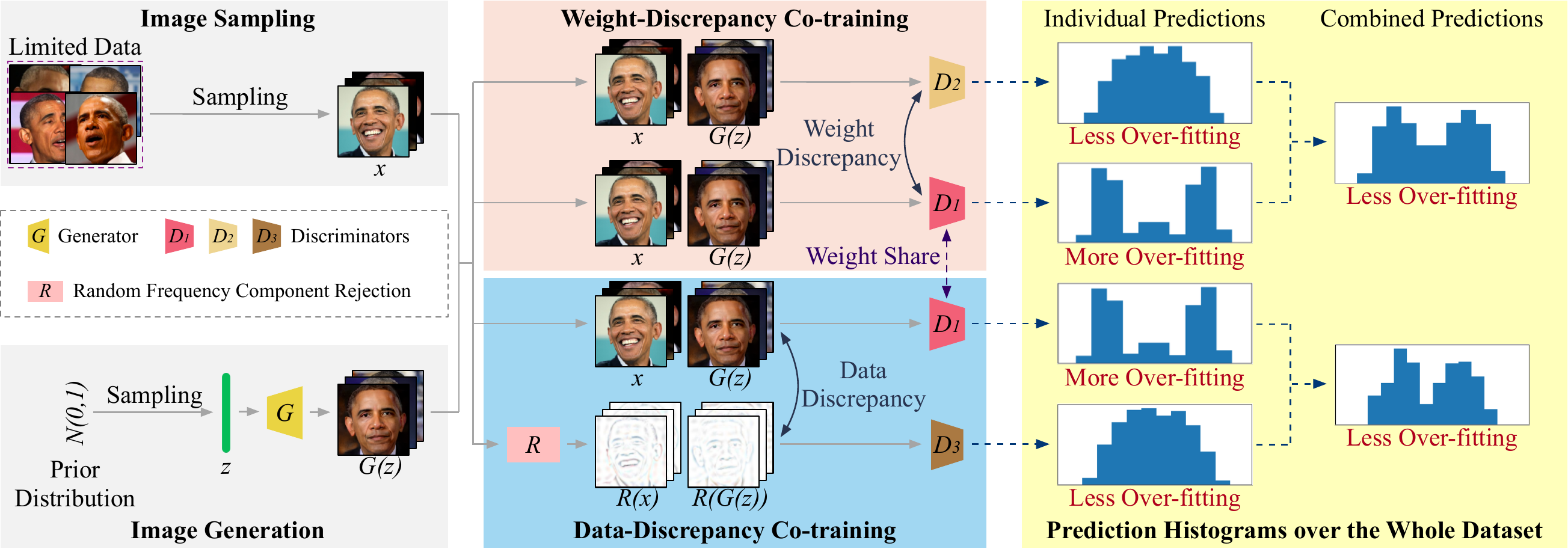}
\caption{
The architecture of the proposed GenCo: GenCo consists of four modules on Image Sampling, Image Generation, Weight-Discrepancy Co-training (WeCo) and Data-Discrepancy Co-training (DaCo). Image Sampling samples images $x$ from limited training data and Image Generation generates images $G(z)$ with a generator $G$. $x$ and $G(z)$ are fed to WeCo to co-train discriminators $D_1$ and $D_2$ which are differentiated by a weight discrepancy loss. They are also fed to DaCo to co-train discriminators $D_1$ and $D_3$, where a different view of $x$ (produced by the Random Frequency Component Rejection module $R$) is fed to $D_3$. 
The box on the right shows six prediction histograms over the whole dataset. The left four are produced by $D_1$ (2 with shared weights), $D_2$ and $D_3$, and the right two are the combined prediction histograms by WeCo and DaCo, respectively. The horizontal axis of these histograms shows the discriminator score in [-4, 4] and the vertical axis shows the numbers of occurrence. The four distinctive yet complementary discriminators capture different information of the training images, and the fusion of them with more comprehensive information mitigates the discriminator over-fitting issue effectively.
}
\label{fig:model}
\end{figure*}

\section{Related Works}

\textbf{Generative Adversarial Networks (GANs):} The pioneer generative adversarial network~\cite{goodfellow2014generative} greatly changes the paradigm of automated image generation. Leveraging the idea in~\citet{goodfellow2014generative}, quite a few GANs have been developed for realistic and high-fidelity image generation in the past few years. They strive to improve the generation realism and fidelity from different aspects by introducing task-specific training objectives~\cite{arjovsky2017wasserstein, gulrajani2017improved, mao2017least}, designing more sophisticated network architectures~\cite{miyato2018spectral, miyato2018cgans, zhang2019self}, and adopting very different training strategies~\cite{karras2017progressive,zhang2017stackgan,liu2020diverse}, etc. On the other hand, most existing GANs still require a large number of training images for capturing the data distributions comprehensively. When only a limited number of training images are available, they often suffer from clear over-fitting in discriminators and their generated images also degrade significantly.

We target data-limited image generation, which strives to learn robust generation models from limited training images yet without sacrificing much generation quality.

\textbf{Data-Limited Image Generation:} Data-limited image generation has attracted increasing interest for mitigating the laborious and time-consuming image collection process. The earlier studies~\cite{webster2019detecting, gulrajani2020towards} suggest that one of the main obstacles of training GANs with limited training data is the discriminator over-fitting issue. 
The recent studies strive to address the issue through massive data augmentation. For example, \citet{zhao2020differentiable} introduces different types of differentiable augmentation to stabilize the network training process which leads to a clear improvement in generation realism and generation fidelity. \citet{karras2020training} presents an adaptive augmentation mechanism that effectively mitigates discriminator over-fitting and undesirable leaking of augmentation to the generated images.

In this paper,  we tackle the discriminator over-fitting issue from a different perspective and propose Generative Co-training that employs the idea of co-training to view the limited data from  multiple complementary views.

\textbf{Co-training:} Co-training aims to learn multiple complementary classifiers from different views for training more generalizable models. The idea traces back to a few pioneer studies~\cite{blum1998combining, sun2011robust, yu2011bayesian} that propose co-training to tackle the data insufficiency problem while training classification models. With the recent advance of deep neural networks and demands for larger amounts of training data, the idea of co-training has attracted increasing interest in various deep network training tasks. For example, \citet{qiao2018deep} presents a deep co-training technique that encourages view differences by training multiple deep neural networks in a semi-supervised image recognition task. \citet{saito2018maximum} adopts the co-training idea to align the feature category between source and target domains.

We introduce co-training into the data-limited image generation task for mitigating its inherent over-fitting issue. To the best of our knowledge, this is the first work that explores the discriminative co-training idea for the generative image generation task. Extensive experiments show its effectiveness, more details to be described in the Experiments.

\section{Method}

This section describes the detailed methodology of the proposed GenCo.  As illustrated in Fig. \ref{fig:model}, we co-train multiple distinctive discriminators to mitigate the over-fitting issue. In addition, we design two instances of GenCo, including a Weight-Discrepancy Co-training (WeCo) that trains multiple distinctive discriminators by diversifying their parameters and a Data-Discrepancy Co-training (DaCo) that trains multiple distinctive discriminators by feeding them with different views of training images. We focus on two discriminators in WeCo and DaCo and will discuss the extension with more discriminators in Experiments. The ensuing subsections will describe the problem definition of data-limited image generation, the network architecture of GenCo, details of the proposed WeCo and DaCo, and the overall training objective, respectively.

\subsection{Problem Definition}
\label{sec:Problem Definition}
The GAN models are the cornerstone techniques for image generation tasks. Each GAN consists of a discriminator $D$ and a generator $G$. The general loss function for discriminator and generator is defined as:

\begin{align}\left.\begin{aligned}
\label{eqn:general D}
\mathcal{L}_{d}(D; x, G(z)) = \mathbb{E}[\log(D(x))] \\ 
+  \mathbb{E}[\log(1-D(G(z))]
\end{aligned}\right.\end{align}
\begin{align}\left.\begin{aligned}
\label{eqn: general G}
\mathcal{L}_{g}(D; G(z)) = \mathbb{E}[\log(1-D(G(z))]
\end{aligned}\right.\end{align}

\noindent where $\mathcal{L}_{d}$ and $\mathcal{L}_{g}$ denote the discriminator and generator losses, respectively. $x$ denotes a training sample and $z$ is sampled from a prior distribution.

\begin{table*}
\begin{center}
\resizebox{0.8\textwidth}{!}{
\begin{tabular}{l|c|c| ccc|cc}
\hline
 \multirow{2}*{{Methods}} & Massive & \multirow{1}*{{Pre-training}}  & \multicolumn{3}{c|}{100-shot}  & \multicolumn{2}{c}{AFHQ}\\
 & Augmentation &w/ 70K images& Obama & Grumpy Cat & Panda & Cat & Dog\\
\hline

Scale/shift~\cite{noguchi2019image}& No & Yes & 50.72 & 34.20 & 21.38 & 54.83 & 83.04\\
MineGAN~\cite{wang2020minegan}& No&Yes& 50.63 &34.54 &14.84& 54.45& 93.03 \\
TransferGAN~\cite{wang2018transferring} & No &Yes &48.73 & 34.06 &23.20 &52.61& 82.38 \\
TransferGAN + DA~\cite{zhao2020differentiable}  & No &Yes &39.85 &29.77& 17.12 &49.10 &65.57\\
FreezeD~\cite{mo2020freeze} & No & Yes & 41.87 & 31.22& 17.95& 47.70& 70.46 \\

\hline
StyleGAN2~\cite{karras2020analyzing} & No & No & 80.20 & 48.90 & 34.27 & 71.71 & 130.19\\
LeCam-GAN~\cite{tseng2021regularizing} & No & No & 38.58 & 41.38 & 19.88 & 60.26 & 112.39\\

\textbf{GenCo}  & No & No & \textbf{36.35} & \textbf{33.57}&  \textbf{15.50} & \textbf{54.78} & \textbf{94.47}\\
\hline
DA~\cite{zhao2020differentiable} & Yes & No &46.87& 27.08& 12.06& 42.44& 58.85 \\

ADA~\cite{karras2020training} & Yes &No & 45.69 & 26.62& 12.90 & 40.77 &56.83\\

LeCam-GAN~\cite{tseng2021regularizing}  & Yes & No & 33.16 & 24.93 & 10.16 & 34.18 & 54.88\\
 
\textbf{GenCo} & Yes & No  &\textbf{32.21} & \textbf{17.79} & \textbf{9.49} &\textbf{30.89} & \textbf{49.63} \\
\hline
\end{tabular}}
\end{center}
\caption{Comparison with the state-of-the-arts over 100-shot and AFHQ: Training with 100 (Obama, Grumpy Cat and Panda), 160 (AFHQ Cat), and 389 (AFHQ Dog) samples, GenCo performs the best consistently. It achieves comparable results as transfer learning methods (Rows 1-5) pre-trained with 70K images. We report FIDs ($\downarrow$) averaged over three runs.}
\label{Low Shot Generation}
\end{table*}

\begin{figure*}[t!] 
\centering
\includegraphics[width=0.8\linewidth]{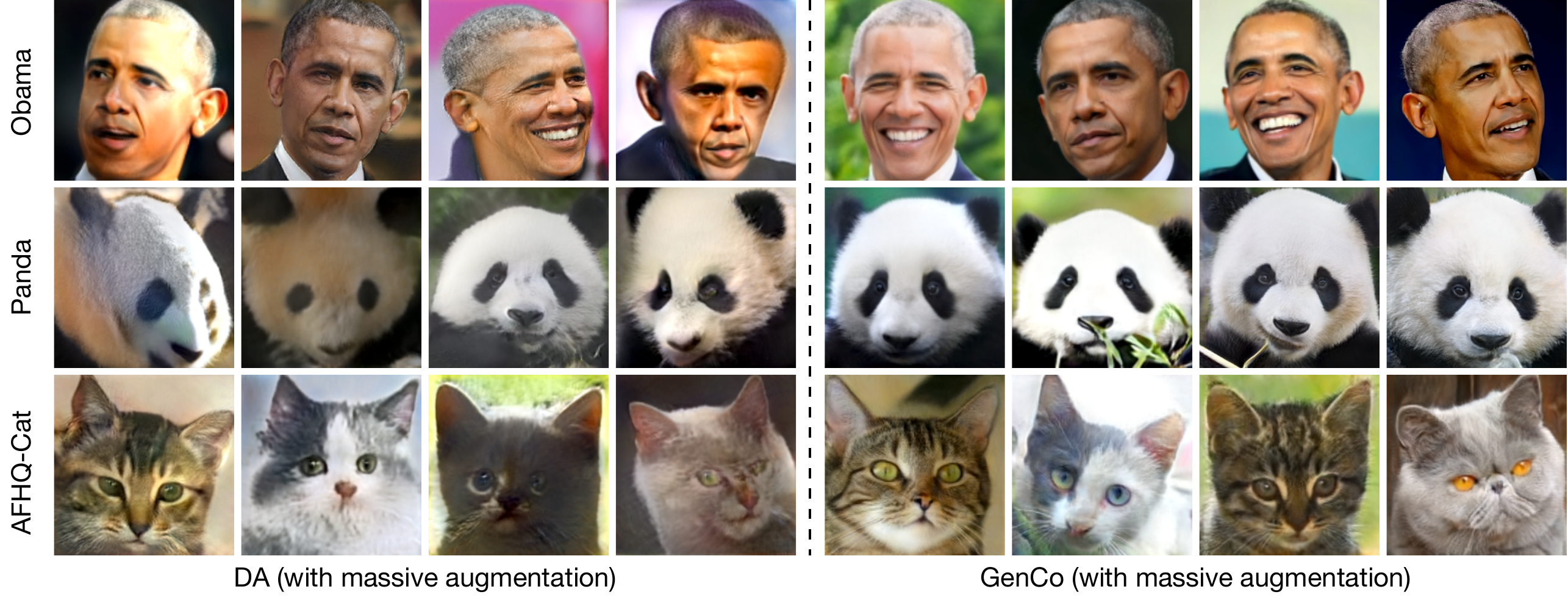}
\caption{
Qualitative results over 100-shot datasets ($e.g.$, Obama and Panda) and AFHQ dataset ($e.g.$, Cat):
The generation by GenCo is clearly more realistic than that by DA~\cite{zhao2020differentiable}, the state-of-the-art data-limited generation method.
} 
\label{fig:qualitative}
\end{figure*}

\begin{table*}
\begin{center}
\setlength{\tabcolsep}{3pt}
\resizebox{0.95\textwidth}{!}{
\begin{tabular}{l|c|ccc|ccc}
\hline
\multirow{2}*{Methods} &Massive& \multicolumn{3}{c|}{CIFAR-10}  & \multicolumn{3}{c}{CIFAR-100}\\
\cline{3-8}
&Augmentation&  \multicolumn{1}{l}{100\% data  }  & \multicolumn{1}{l}{20\% data}   & \multicolumn{1}{l|}{10\% data}  & \multicolumn{1}{l}{100\% data}  & \multicolumn{1}{l}{20\% data }  & \multicolumn{1}{l}{10\% data}  \\
\hline
Non-saturated GAN~\cite{goodfellow2014generative}& No &9.83\scriptsize$\pm$0.06  &18.59\scriptsize$\pm$0.15   &41.99\scriptsize$\pm$0.18   &13.87\scriptsize$\pm$0.08 & 32.64 \scriptsize$\pm$0.19 & 70.5\scriptsize$\pm$0.38\\
LS-GAN~\cite{mao2017least}& No &9.07\scriptsize$\pm$0.01 & 21.60\scriptsize$\pm$0.11 & 41.68\scriptsize$\pm$0.18 & 12.43\scriptsize$\pm$0.11 & 27.09\scriptsize$\pm$0.09 & 54.69\scriptsize$\pm$0.12  \\
RAHinge GAN~\cite{jolicoeur2018relativistic}
& No & 11.31\scriptsize$\pm$0.04 & 23.90\scriptsize$\pm$0.22& 48.13\scriptsize$\pm$0.33& 14.61\scriptsize$\pm$0.21& 28.79\scriptsize$\pm$0.17 & 52.72\scriptsize$\pm$0.18\\
StyleGAN2~\cite{karras2020analyzing} & No & 11.07\scriptsize$\pm$0.03  & 23.08\scriptsize$\pm$0.11  & 36.02\scriptsize$\pm$0.15  &16.54\scriptsize$\pm$0.04  & 32.30 \scriptsize$\pm$0.11 & 45.87\scriptsize$\pm$0.15 \\
BigGAN~\cite{brock2018large} & No & 9.07\scriptsize$\pm$0.06& 21.86\scriptsize$\pm$0.29& 48.08\scriptsize$\pm$0.10 & 13.60\scriptsize$\pm$0.07 & 32.99\scriptsize$\pm$0.24 &66.71\scriptsize$\pm$0.01\\
\textbf{GenCo} & No & \textbf{8.83}\scriptsize$\pm$0.04 & \textbf{16.57}\scriptsize$\pm$0.08 &  \textbf{28.08}\scriptsize$\pm$0.11&  \textbf{11.90}\scriptsize$\pm$0.02& \textbf{26.15}\scriptsize$\pm$0.08& \textbf{40.98}\scriptsize$\pm$0.09\\\hline
DA~\cite{zhao2020differentiable}   & Yes & 8.75\scriptsize$\pm$0.03 &  14.53\scriptsize$\pm$0.10& 23.34\scriptsize$\pm$0.09 & 11.99\scriptsize$\pm$0.02&22.55\scriptsize$\pm$0.06&35.39\scriptsize$\pm$0.08 \\
\textbf{GenCo} & Yes &\textbf{7.98}\scriptsize$\pm$0.02 & \textbf{12.61}\scriptsize$\pm$0.05 & \textbf{18.10}\scriptsize$\pm$0.06 & \textbf{10.92}\scriptsize$\pm$0.02 &\textbf{18.44}\scriptsize$\pm$0.04&\textbf{25.22}\scriptsize$\pm$0.06 \\
\hline
\end{tabular}
}
\end{center}
\caption{
Comparing GenCo with the state-of-the-arts over CIFAR: GenCo mitigates the discriminator over-fitting issue and outperforms the state-of-the-arts consistently. We report FID ($\downarrow$) scores averaged over three runs. Note GenCo and DA~\cite{zhao2020differentiable} are implemented on BigGAN framework in this experiment.}
\label{CIFAR-quantative}
\end{table*}

With limited training data $X_{L}$,
discriminator in GANs tends to become over-fitting, leading to sub-optimal image generation. Concretely, the over-fitting discriminator produces high prediction scores and very small discriminator loss $\mathcal{L}_{d}$. The very small discriminator loss leads to very large generator loss $\mathcal{L}_{g}$ as well as gradients which accumulate during training and further cause training divergence and degraded generation. The following subsections describe how the proposed GenCo mitigates the discriminator over-fitting issue.

\subsection{Overview of Network Architecture}  \label{sec:Overview of Network Architectur}

GenCo consists of four major modules as demonstrated in Fig. \ref{fig:model}: Image Sampling, Image Generation,  Weight-Discrepancy Co-training and Data-Discrepancy Co-training. 
Image Sampling samples images $x$ from the limited dataset $X_{L}$ and Image Generation generates fake samples $G(z)$ from a prior distribution with generator $G$.
$x$ and $G(z)$ are then passed to WeCo to co-train discriminators $D_1$ and $D_2$ that are differentiated by a weight discrepancy loss as defined in Eqs.\ref{eqn:D1} and \ref{eqn:D2}. Meanwhile, $x$ and $G(z)$ are also fed to DaCo to co-train discriminators $D_1$ and $D_3$ that are differentiated by distinctive views of the inputs as defined in Eqs.\ref{eqn:D1daco} and \ref{eqn:D3}, where $D_1$ is fed with the original $x$ while $D_3$ is fed with partial frequency components of $x$ (generated by the proposed Random Frequency Component Rejection ($R$)).

\subsection{Weight-Discrepancy Co-training} \label{sec:weight-discrepancy co-training}

The proposed WeCo aims to learn two distinctive discriminators $D_1$ and $D_2$ by diversifying their parameters. We achieve diverse parameters by defining a weight discrepancy loss $\mathcal{L}_{wd}$ that minimizes the cosine distance between the weights of $D_1$ and $D_2$:
\begin{align}\begin{aligned}
\label{equ:weight}
\mathcal{L}_{wd}(D_1, D_2)= \frac{\overrightarrow{W_{D_{1}}}\overrightarrow{W_{D_{2}}}}{|\overrightarrow{W_{D_{1}}}||\overrightarrow{W_{D_{2}}}|}
\end{aligned}\end{align}
where $\overrightarrow{W_{D_{1}}}$ and $\overrightarrow{W_{D_{2}}}$ are the weights of $D_1$ and $D_2$. The loss of $D_1$ and $D_2$ can thus be formulated by:
\begin{align}\left.\begin{aligned}
\label{eqn:D1}
\mathcal{L}_{D_1} = \mathcal{L}_{d}(D_1; x, G(z)) 
\end{aligned}\right.\end{align}
\begin{align}\left.\begin{aligned}
\label{eqn:D2}
\mathcal{L}_{D_2} = \mathcal{L}_{d}(D_2; x, G(z))  +
\mathcal{L}_{wd}(D_1, D_2)
\end{aligned}\right.\end{align}
where $L_d$ is the general discriminator loss as in Eq.\ref{eqn:general D}.  
$\mathcal{L}_{wd}$ is the weight discrepancy loss as defined in Eq.\ref{equ:weight}. We apply $\mathcal{L}_{wd}$ on only one discriminator for simplicity because applying it on two discriminators does not make much difference. 

The overall WeCo loss $\mathcal{L}_{D_1,D_2}^{WeCo}$ can thus be defined by:  
\begin{align}\left.\begin{aligned}
\label{eqn:weco}
\mathcal{L}_{D_1,D_2}^{WeCo} = \mathcal{L}_{D_1} + \mathcal{L}_{D_2}
\end{aligned}\right.\end{align}

 \subsection{Data-Discrepancy Co-training} \label{sec:data-discrepancy co-training}
DaCo co-trains two distinctive discriminators $D_1$ and $D_3$ that take different views of the input images. Specifically,  $D_1$ is fed with the original images while $D_3$ takes partial frequency components (FCs) of the input images (generated by Random Frequency Component Rejection ($R$)) as input.

The component $R$ consists of three processes including $R_t$, $R_r$, and $R_{t^{-1}}$. Specifically, $R_t$ first converts the images $x$ from spatial space to frequency space. $R_r$ then rejects certain FCs randomly with the rest FCs intact. Finally, $R_{t^{-1}}$ converts the intact FCs back to spatial space to form the new inputs of $D_3$. Detailed definitions of $R_t, R_r, R_{t^{-1}}$ are available in the supplementary material. Note the percentage of the rejected FCs is controlled by a hyper-parameter $P$ which does not affect the generation much as shown in Table \ref{control varibla p}. We empirically set $P$ at 0.2 in our network.

The loss functions of $D_1$ 
and $D_3$ 
can thus be defined by: 
\begin{align}\left.
\begin{aligned}
\label{eqn:D1daco}
\mathcal{L}_{D_1} = \mathcal{L}_{d}(D_1; x, G(z)) 
\end{aligned}
\right.\end{align}
\begin{align}\left.\begin{aligned}
\label{eqn:D3}
\mathcal{L}_{D_3} = \mathcal{L}_{d}(D_3; R(x), R(G(z)))
\end{aligned}\right.\end{align}

\noindent where the loss of $D_1$ is the same as the loss of $D_1$ (Eq. \ref{eqn:D1}) in WeCo (they share weights). The loss of $D_3$ is close to that of $D_1$ and the differences are largely due to the different inputs by the Random Frequency Component Rejection ($R$).

The overall DaCo loss $\mathcal{L}_{D_1,D_3}^{DaCo}$ can thus be defined by:
\begin{align}\left.\begin{aligned}
\label{eqn:daco}
\mathcal{L}_{D_1,D_3}^{DaCo} = \mathcal{L}_{D_1} + \mathcal{L}_{D_3}
\end{aligned}\right.\end{align}

\begin{table}[t]
\centering
\setlength{\tabcolsep}{3pt}
\resizebox{1\columnwidth}{!}{
\begin{tabular}{l|cccc|cccc}
\hline
\multirow{2}*{{Methods}}  & \multicolumn{4}{c|}{FFHQ} & \multicolumn{4}{c}{LSUN-Cat} \\
& 30K&10K & 5K&1K &30K& 10K & 5K&1K \\
\hline

StyleGAN2  &11.22& 27.56 & 42.32& 92.86 & 14.28& 46.98& 90.12 & 178.31\\
\textbf{GenCo} &\textbf{8.27}   &\textbf{15.66}  & \textbf{27.96}  & \textbf{65.31} &\textbf{12.25} & \textbf{20.15} &\textbf{ 40.79} & \textbf{140.08}
\\
\hline
\end{tabular}}
\caption{Quantitative results on the FFHQ and LSUN-Cat datasets
: We report FID ($\downarrow$) over three runs.}
\label{tab:FFHQ}
\end{table}

\subsection{Overall Training Objective} \label{sec:Overall Training Objective}

The generator $G$ learns with information from all three discriminators. Its loss $\mathcal{L}_{G}^{total}$ can be formulated by:
\begin{align}\left.\begin{aligned}
\mathcal{L}_{G}^{total}= \mathcal{L}_{g}(D_1; G(z)) + \mathcal{L}_{g}(D_2; G(z)) \\ + \mathcal{L}_{g}(D_3; R(G(z)))
\end{aligned}\right.\end{align}

The overall training objective of the proposed GenCo can thus be formulated by,

\begin{align}\left.\begin{aligned}
\mathop{min}\limits_{G}\mathop{max}\limits_{D_1, D_2, D_3}  \mathcal{L}_{G}^{total} + \mathcal{L}_{D_1,D_2}^{WeCo} + \mathcal{L}_{D_1,D_3}^{DaCo}
\end{aligned}\right.\end{align}

\textbf{Why is GenCo effective?}
In data-limited image generation, one major issue is that discriminator in GANs tends to suffer from over-fitting by capturing certain simple structures and patterns only~\cite{bau2019seeing, zhang2021defect}. The proposed GenCo mitigates this issue by co-training two discriminators in WeCo and DaCo. 
With the co-training design, although one discriminator ($e.g.$, $D_1$) may overfit and focuses on learning simple structures and patterns, the other distinctive discriminator ($e.g.$, $D_2$ in WeCo and $D_3$ in DaCo) with different parameters or data inputs will be encouraged to learn different information like complex structures and patterns. The two discriminators thus complement each other to focus on different types of information, which helps mitigate the discriminator over-fitting issue effectively (as shown in Fig.\ref{fig:model}). From another view, the intrinsic cause of the discriminator over-fitting is the large generator loss that leads to training divergence. In GenCo, the overall over-fitting with two distinctive discriminators in either WeCo or DaCo is reduced which leads to smaller generator loss and further mitigates training divergence.

In addition, WeCo and DaCo in GenCo also complement each other to mitigate the overall over-fitting as they achieve co-training from different perspectives. Specifically, WeCo achieves co-training by diversifying the discriminator parameters, whereas DaCo achieves co-training by feeding two discriminators with different views of the inputs.

\section{Experiments}
In this section, we conduct extensive experiments to evaluate our proposed GenCo. We first briefly introduce the datasets and evaluation metrics used in our experiments (section \ref{sec:dataset}).
We then benchmark GenCo across these datasets and provide a visualization of GenCo
(section \ref{sec:cifar}, \ref{sec:lowshot}, \ref{sec:ffhq}, \ref{sec:activation map}).
Moreover, we conduct extensive ablation studies
(section \ref{sec:abalation}) 
and discussions 
(section \ref{sec:discussion}) 
to support our design choices. 
All the experiments are based on StyleGAN2 framework unless specified otherwise.

\subsection{Datasets and Evaluation Metrics} \label{sec:dataset}
We conduct experiments over  multiple public datasets: CIFAR (Krizhevsky et al.  2009), 100-shot~\cite{zhao2020differentiable},  AFHQ~\cite{si2011learning}, FFHQ~\cite{karras2019style}
and LSUN-Cat~\cite{yu2015lsun}. 
We follow ~\citet{zhao2020differentiable} and perform evaluations with two widely adopted metrics in image generation: Frechet Inception Distance (FID)~\cite{heusel2017gans} and inception score (IS)~\cite{salimans2016improved}.
The validation set is used for FID calculation for CIFAR. The full training set is used for FID calculation for 100-shot, AFHQ, FFHQ
and LSUN-Cat.

\subsection{Experments on 100-shot and AFHQ} \label{sec:lowshot}

The bottom part of Table \ref{Low Shot Generation} compares GenCo with state-of-the-art methods in data-limited image generation ($i.e.$, DA, ADA and LeCam-GAN) over 100-shot and AFHQ. We can see that GenCo performs the best consistently, demonstrating the effectiveness of GenCo in mitigating discriminator over-fitting.

Table \ref{Low Shot Generation} (Rows 6 and 8) 
compares GenCo with state-of-the-art GANs ($i.e.$, StyleGAN2).
It shows that GenCo improves the generation consistently by large margins. In addition, several studies explore transfer learning by pre-training the model with large datasets. The top part of Table \ref{Low Shot Generation} shows their FID scores (pre-trained with FFHQ with 70K images). We can see that GenCo achieves comparable FIDs by using only 100 – 400 training samples instead. Fig. \ref{fig:qualitative} qualitatively demonstrates that GenCo outperforms the state-of-the-art in data-limited generation, especially in terms of the generated shapes and textures.

\subsection{Experiments on CIFAR-10 and CIFAR-100}  \label{sec:cifar}

We compare GenCo with DA~\cite{zhao2020differentiable}, the state-of-the-art in data-limited generation at the bottom of Table~\ref{CIFAR-quantative}. 
It shows that GenCo outperforms DA consistently under the massive augmentation setup, demonstrating the effectiveness of our GenCo in mitigating the discriminator over-fitting.

Table \ref{CIFAR-quantative} (Rows 1-6) also quantitatively compares GenCo with several state-of-the-art GANs over datasets CIFAR-10 and CIFAR-100.
We can see that GenCo performs the best consistently especially when training samples are limited.
The superior performance is largely attributed to the co-training idea in GenCo which mitigates the discriminator over-fitting effectively.
Evaluations in IS are provided in the supplementary material.

\begin{figure}[t] 
\centering
 \includegraphics[width=0.95\linewidth]{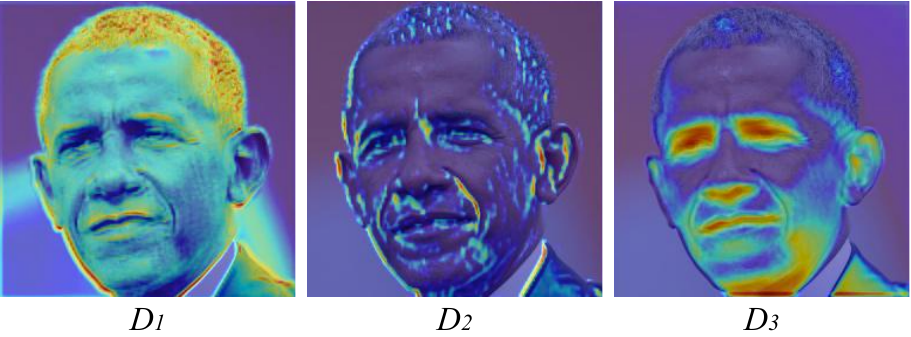}
\caption{
Activation maps of discriminators in GenCo: GenCo mitigates discriminator over-fitting with three distinctive discriminators that capture complementary information. As illustrated, $D_1$, $D_2$, and $D_3$ attend to facial styles (color, brightness, etc.), facial details (wrinkles, face outline, etc.) and facial expressions 
(eyes, mouth, etc.), respectively.
}
\label{fig:Visualization}
\end{figure}

\renewcommand\arraystretch{1.1}
\begin{table}[t]
\centering
\resizebox{0.9\columnwidth}{!}{
\begin{tabular}{ cc| c|c}
\hline
  \multicolumn{2}{c|}{Design Choice} & \multicolumn{1}{c|}{Cifar-10} & 100shot\\
\hline
WeCo & DaCo & 10\% data & Obama\\
\hline
-&- & 48.08\scriptsize$\pm$0.10 & 80.16\scriptsize$\pm$0.22\\
$\checkmark$&-& 34.05\scriptsize$\pm$0.15  & 55.34\scriptsize$\pm$0.17\\
-  & $\checkmark$ & 30.33\scriptsize$\pm$0.13& 41.96\scriptsize$\pm$0.19\\

$\checkmark$& $\checkmark$&\textbf{28.08}\scriptsize$\pm${0.11} &\textbf{36.28}\scriptsize$\pm${0.11}\\
\hline
\end{tabular}
}
\caption{
Ablation study of GenCo: WeCo and DaCo in GenCo both mitigate discriminator over-fitting effectively with improved generation over the baseline. GenCo performs simply the best as WeCo and DaCo are complementary to each other. The FIDs ($\downarrow$) are averaged over three runs.
}
\label{tab:ablation}
\end{table}

\begin{figure}[t] 
\centering
 \includegraphics[width=0.95\linewidth]{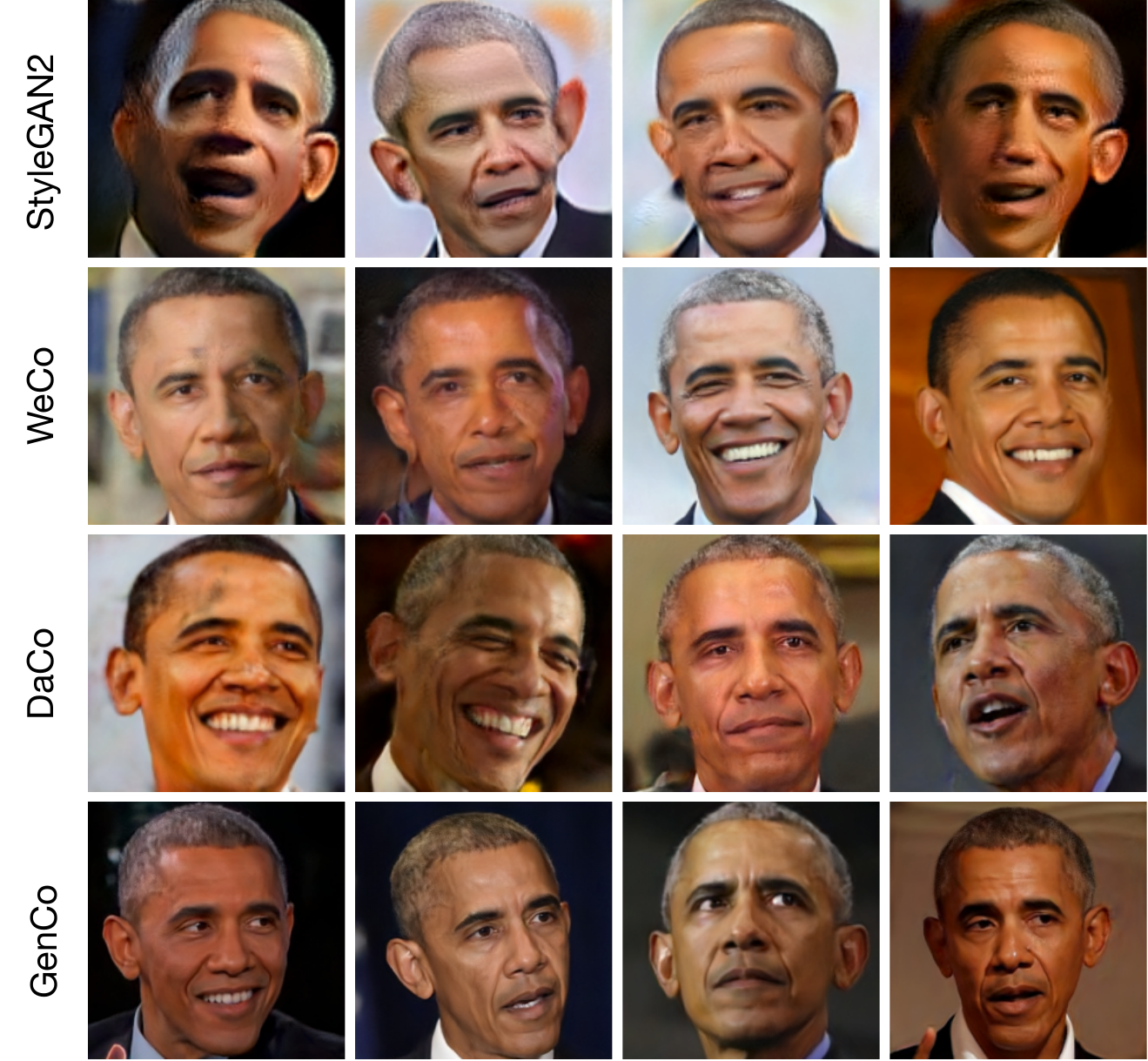}
   \caption{
   Qualitative ablation study over 100-shot Obama: The generation by WeCo (Row 2) and DaCo (Row 3) alone is clearly more realistic than the generation by the baseline (Row 1). In addition, the generation by GenCo (Row 4) that combines WeCo and GenCo is most realistic.
   }
\label{abalation-qualitative}
\end{figure}

\renewcommand\arraystretch{1.1}
\begin{table}[t]
\setlength{\tabcolsep}{2.5pt}
\centering
\resizebox{1\columnwidth}{!}{
\begin{tabular}{l|cccc}
\hline
 \multirow{2}*{{Methods}} & \multicolumn{2}{c}{Baseline} &\multicolumn{2}{c}{\textbf{+GenCo}} \\
&FID ($\downarrow$)&IS ($\uparrow$)&FID ($\downarrow$)&IS ($\uparrow$)\\
\hline
BigGAN & 48.08\scriptsize$\pm$0.10&7.09\scriptsize$\pm$0.03&\textbf{28.08}\scriptsize$\pm$0.11 &\textbf{8.01}\scriptsize$\pm$0.26\\
+ noise~\cite{sonderby2016amortised} & 47.06\scriptsize$\pm$0.11&7.12\scriptsize$\pm$0.05&\textbf{27.88}\scriptsize$\pm$0.11 &\textbf{8.06}\scriptsize$\pm$0.12\\
+ CR~\cite{zhang2019consistency}& 44.16\scriptsize$\pm$0.10& 7.27\scriptsize$\pm$0.04& \textbf{27.03}\scriptsize$\pm$0.08 &  \textbf{8.12}\scriptsize$\pm$0.11\\
+ GP-0 (Mescheder et al. 2018)& 42.22\scriptsize$\pm$0.18& 7.38\scriptsize$\pm$0.03& \textbf{26.58}\scriptsize$\pm$0.12& \textbf{8.15}\scriptsize$\pm$0.06\\
+ LeCam-GAN~\cite{tseng2021regularizing}&35.23\scriptsize$\pm$0.14&7.97\scriptsize$\pm$0.03&\textbf{25.89}\scriptsize$\pm$0.07& \textbf{8.23}\scriptsize$\pm$0.25\\\hline 
\end{tabular}
}
\caption{
Experiments on GenCo and regularization-based generation methods: GenCo and regularization-based methods are clearly complementary in data-limited generation.  
The FIDs ($\downarrow$) and IS  ($\uparrow$) are averaged over three runs.
}
\label{tab:regularization}
\end{table}

\renewcommand\arraystretch{1.1}
\begin{table}[t]
\setlength{\tabcolsep}{3pt}
\centering
\resizebox{0.95\columnwidth}{!}{
\begin{tabular}{l|cc}
\hline
 \multirow{1}*{{Methods}} & \multicolumn{1}{c}{Baseline} &\multicolumn{1}{c}{\textbf{+GenCo}} \\
BigGAN~\cite{brock2018large}&48.08\scriptsize$\pm$0.10&\textbf{28.08}\scriptsize$\pm$0.11 \\
LS-GAN~\cite{mao2017least}&41.68\scriptsize$\pm$0.18&\textbf{26.64}\scriptsize$\pm$0.15\\
RAHinge GAN~
(Jolico-Martin 2018)
& 48.13\scriptsize$\pm$0.33 & \textbf{36.47}\scriptsize$\pm$0.23 \\
BigGAN + DA~\cite{zhao2020differentiable}&23.34\scriptsize$\pm$0.28& \textbf{18.10}\scriptsize$\pm$0.13\\
\hline
\end{tabular}
}
\caption{
Experiments on the generalization of GenCo with different baselines (FIDs ($\downarrow$) averaged over three runs).
}
\label{tab:baseline}
\end{table}

\subsection{Experiments on FFHQ and LSUN-Cat} \label{sec:ffhq}

Table \ref{tab:FFHQ}  quantitatively compares GenCo with StyleGAN2 over FFHQ and LSUN-Cat. Following DA~\cite{zhao2020differentiable}, we evaluate on 30K, 10K, 5K and 1K training samples. As Table \ref{tab:FFHQ} shows, GenCo improves the baseline consistently.
Note that experiments over FFHQ and LSUN-Cat are trained with 8 GPUs with a maximum training length of 25M images; we thus compare GenCo with the representative StyleGAN2 only due to resource limitations.

\subsection{Visualization of GenCo}
\label{sec:activation map}

GenCo mitigates the discriminator over-fitting effectively by co-training multiple distinctive discriminators ($D_1$ and $D_2 $ in WeCo, $D_1$ and $D_3$ in DaCo) that learn from different views and capture complementary information. This can be observed from their activation maps~\cite{selvaraju2017grad} in Fig. \ref{fig:Visualization} which show that the three discriminators attend and capture different types of visual information. The fusion of them thus provides more comprehensive supervision signals which lead to less discriminator over-fitting, stabler training, and finally better image generation.

\subsection{Ablation study} \label{sec:abalation}
The proposed GenCo consists of two major components, namely, WeCo and DaCo. We study the two components separately to examine their contributions to the overall generation performance. As Table \ref{tab:ablation} shows, including either WeCo or DaCo outperforms the baseline clearly, demonstrating the effectiveness of the proposed co-training idea which mitigates discriminator over-fitting by learning from multiple distinctive views. In addition, combining WeCo and DaCo performs clearly the best which verifies that the distinctive views in WeCo (by weight discrepancy) and DaCo (by input discrepancy) are complementary to each other.

Qualitative ablation studies in Fig. \ref{abalation-qualitative} show that the proposed WeCo and DaCo can produce clearly more realistic generation than baseline, demonstrating the effectiveness of the proposed co-training idea. In addition, GenCo produces the most realistic generation, which verifies that WeCo and DaCo complement each other.

\renewcommand\arraystretch{1.1}
\begin{table}[t]
\centering
\resizebox{0.9\columnwidth}{!}{
\begin{tabular}{cccccc}
\hline
Metrics & Baseline & $R$ as augmentation& DaCo
\\\hline
FID ($\downarrow$)&48.08\scriptsize$\pm$0.10 & 40.36\scriptsize$\pm$0.11&\textbf{30.33}\scriptsize$\pm$0.13 \\
IS ($\uparrow$) &7.09\scriptsize$\pm$0.03 & \textbf{7.43}\scriptsize$\pm$0.18 &\textbf{7.85}\scriptsize$\pm$0.21\\
\hline
\end{tabular}
}
\caption{
Experiments on the random frequency component rejection $R$ in Daco (results averaged over three runs).
}
\label{tab:RFA as Aug}
\end{table}

\renewcommand\arraystretch{1.1}
\begin{table}[t]
\setlength{\tabcolsep}{2.5pt}
\centering
\resizebox{\columnwidth}{!}{
\begin{tabular}{cccccccc}
\hline
 \multirow{2}*{{Metrics}} & \multicolumn{5}{c}{Percentage of rejected frequency components}
\\
 & \multicolumn{1}{c}{0.1} & \multicolumn{1}{c}{0.2} & \multicolumn{1}{c}{0.3} & \multicolumn{1}{c}{0.4} & \multicolumn{1}{c}{0.5}
\\\hline

FID ($\downarrow$) &33.02\scriptsize$\pm$0.09&\textbf{28.08}\scriptsize$\pm$0.11& 30.19\scriptsize$\pm$0.13& 31.78\scriptsize$\pm$0.10& 32.77\scriptsize$\pm$0.12\\
IS ($\uparrow$) &7.76\scriptsize$\pm$0.20&\textbf{8.01}\scriptsize$\pm$0.26&7.94\scriptsize$\pm$026 &7.83\scriptsize$\pm$0.17&7.88\scriptsize$\pm$0.18\\
\hline
\end{tabular}
}
\caption{
Experiments on the amount of rejected frequency components in DaCo (results averaged over three runs).
}
\label{control varibla p}
\end{table}

\subsection{Discussion} \label{sec:discussion}
In this subsection, we analyze our GenCo from several perspectives, where all the experiments are based on the CIFAR-10 dataset with 10\% data unless specified otherwise.

\textbf{Complementary with regularization methods:} Existing regularization methods introduce a regularization term to network parameters or training losses to improve training stability and mitigate the discriminator over-fitting issue in data-limited image generation.
The proposed GenCo addresses the same issue from a very different co-training perspective instead, which can complement these regularization approaches effectively. Table \ref{tab:regularization} reveals 
that existing regularization methods do improve the generation clearly. Meanwhile, incorporating GenCo into them further improves the generation consistently by large margins.

\textbf{Generalization of GenCo:} 
The proposed GenCo can work with various baselines with similar performance gains. Table \ref{tab:baseline} shows
that GenCo improves the generation consistently while working with different baselines. 
The superior generalization 
is largely attributed to the co-training design in GenCo, which is independent of the network architectures and training losses.

\textbf{Effectiveness of DaCo:} 
DaCo performs certain light data augmentation as $R$ produces a new input for each input image. 
To demonstrate that DaCo works due to our co-training design instead of the light augmentation, we compare DaCo and its variant that employs $R$ for augmentation only without co-training
Table \ref{tab:RFA as Aug} shows
that DaCo achieves clearly better generation than employing $R$ for augmentation only. This is largely because DaCo employs two distinctive views of the inputs to co-train two different discriminators to mitigate their over-fitting whereas the light augmentation alone does not expand the data distribution much.

\textbf{Robustness of DaCo:} We introduce a hyper-parameter $P$ in DaCo to control the percentage of rejected frequency components (FCs). We perform experiments to study how different $P$ affect the generation performance. As shown in Table \ref{control varibla p}, different $P$ produce quite similar FID and IS. We conjecture that the random rejection of different FCs in each input creates sufficient distinctive views which makes $P$ not that sensitive to the overall generation performance. 

Due to the space limit, we provide more details about the definition of Random Frequency Component Rejection ($R$), description of datasets, and implementations in the supplementary material. In addition, we also provide more quantitative and qualitative experimental results and a thorough complementary study with the state-of-the-art augmentation methods~\cite{zhao2020differentiable, karras2020training} in the supplementary material.

\section{Conclusion}
This paper presents a novel Generative Co-training (GenCo) network that adapts the co-training idea into data-limited generation for tackling its inherent over-fitting issue. We propose two instances of GenCo, namely, Weight-Discrepancy Co-training (WeCo) and Data-Discrepancy Co-training (DaCo). WeCo co-trains multiple distinctive discriminators by diversifying their parameters with a weight discrepancy loss. DaCo achieves co-training by feeding two discriminators with different views of the inputs. We demonstrate that both instances can improve the generation performance and combining WeCo and DaCo achieves the best results. We also show that our GenCo complements state-of-the-art data-augmentation and regularization methods and consistently improves the generation performance.

\newpage

{
\bibliography{aaai22}
}

\end{document}